\documentclass[conference]{IEEEtran}
\IEEEoverridecommandlockouts

\usepackage{balance}
\usepackage{float}
\usepackage[utf8]{inputenc}
\usepackage{graphicx}
\usepackage{amsmath}
\usepackage{makecell}
\usepackage[table]{xcolor}
\def\BibTeX{{\rm B\kern-.05em{\sc i\kern-.025em b}\kern-.08em
    T\kern-.1667em\lower.7ex\hbox{E}\kern-.125emX}}
\PassOptionsToPackage{hyphens}{url}\usepackage{hyperref}

\usepackage{gensymb}

\usepackage{lipsum}
\usepackage[framemethod=tikz]{mdframed}

\usepackage{selinput}
\SelectInputMappings{
adieresis={ä},
germandbls={ß},
}
\usepackage[T1]{fontenc}

\let\labelindent\relax
\usepackage{enumitem}
\usepackage{tikz, xparse}
\usepackage{multirow}
\usepackage{hyperref}
\hypersetup{hidelinks}
\usepackage[newfloat,frozencache=true,cachedir=minted-cache]{minted}

\usepackage{graphicx}
\usepackage[caption=false, font=normalsize, labelfont=sf, textfont=sf]{subfig}

\SetupFloatingEnvironment{listing}{name=Source Code}

\usepackage{tikz} 
\usepackage{epstopdf}
\newcommand{\orcidicon}{\includegraphics[width=0.32cm]{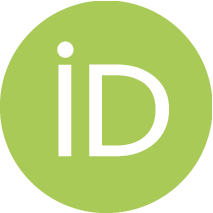}}

\foreach \x in {A, ..., Z}{%
\expandafter\xdef\csname orcid\x\endcsname{\noexpand\href{https://orcid.org/\csname orcidauthor\x\endcsname}{\noexpand\orcidicon}}
}

\usetikzlibrary{decorations.pathreplacing,}

\NewDocumentCommand\mybrace{mmo}{%
\IfValueTF {#3}{%
\begin{tikzpicture}[overlay, remember picture,decoration={brace,amplitude=1ex}]
  \draw[decorate,thick] (#1.north east) -- (#2.south east) node[midway, right=0.1cm] {$=$}node[midway, right=0.5cm,text=black,text width = 2in,] {{#3}};
\end{tikzpicture}%
}%
{%
\begin{tikzpicture}[overlay, remember picture,decoration={brace,amplitude=1ex}]
  \draw[decorate,thick] (#1.north east) -- (#2.south east);
\end{tikzpicture}%
}%
}%

\usepackage{geometry}
\usepackage{etoolbox}
\makeatletter
\patchcmd{\@makecaption}
  {\scshape}
  {}
  {}
  {}
\makeatother

\title{\LARGE \bf
Assessment of the Utilization of Quadruped Robots in Pharmaceutical Research and Development Laboratories}

\author{Brian Parkinson$^{1}$, Ádám Wolf$^{1,2}$, Péter Galambos$^{3}$, Károly Széll$^{4}$
\thanks{$^{1}$ Baxalta Innovations GmbH, a Takeda company, Industriestraße 67, A-1221 Wien, Austria
        {\tt\small (\{brian.parkinson, adam.wolf\}@takeda.com})}%
\thanks{$^{2}$ Doctoral School of Applied Informatics and Applied Mathematics, {\'O}buda University}%
\thanks{$^{3}$Antal Bejczy Center for Intelligent Robotics, {\'O}buda University
        {\tt\small (peter.galambos@irob.uni-obuda.hu)}}%
\thanks{$^{4}$Alba Regia Technical Faculty, {\'O}buda University, H-8000 Sz{\'e}kesfeh{\'e}rv{\'a}r, Hungary
        {\tt\small (szell.karoly@uni-obuda.hu)}}%
}

\begin{document}

\maketitle
\thispagestyle{empty}
\pagestyle{empty}

\begin{abstract}
Drug development is becoming more and more complex and resource-intensive. To reduce the costs and the time-to-market, the pharmaceutical industry employs cutting-edge automation solutions. Supportive robotics technologies, such as stationary and mobile manipulators, exist in various laboratory settings. However, they still lack the mobility and dexterity to navigate and operate in human-centered environments. We evaluate the feasibility of quadruped robots for the specific use case of remote inspection, utilizing the out-of-the-box capabilities of Boston Dynamics' Spot platform. We also provide an outlook on the newest technological advancements and the future applications these are anticipated to enable.

\end{abstract}


\section{Introduction}  \label{sec:intro} 
%

Over the past few decades, drug discovery and development have become increasingly complex in nature \cite{Carney2012Drug2012}. New modalities require a multidisciplinary approach to overcome this hurdle. An example is the adoption of new and innovative lab automation technologies in an effort to increase efficiency and accelerate development. R\&D labs present a highly dynamic and compact environment whereby human operators are required to undertake repetitive and mundane tasks. Examples include the preparation and transportation of standards, reagents, and samples, as well as the maintenance of various unit operations, to name a few.

Advanced and continuous processing is another example of an innovative approach, aiming to accelerate drug production in a more dynamic and flexible manner, which presents many challenges \cite{Domokos2021IntegratedReview}.
One of which is the need for operators to perform an on-call duty to monitor the overall system, check the status of unit operations, and potentially intervene in cases where situations go awry. Recent advancement in the technological readiness of collaborative robotics has revolutionized the pharmaceutical industry's ability to outsource a few of the aforementioned tasks. The adoption of robots is an effort to improve work conditions and allow human operators to undertake more intellectually challenging tasks such as data analysis and the development of novel methods and processes.

\begin{figure}[ht]
\centering
    \includegraphics[width=0.8\linewidth]{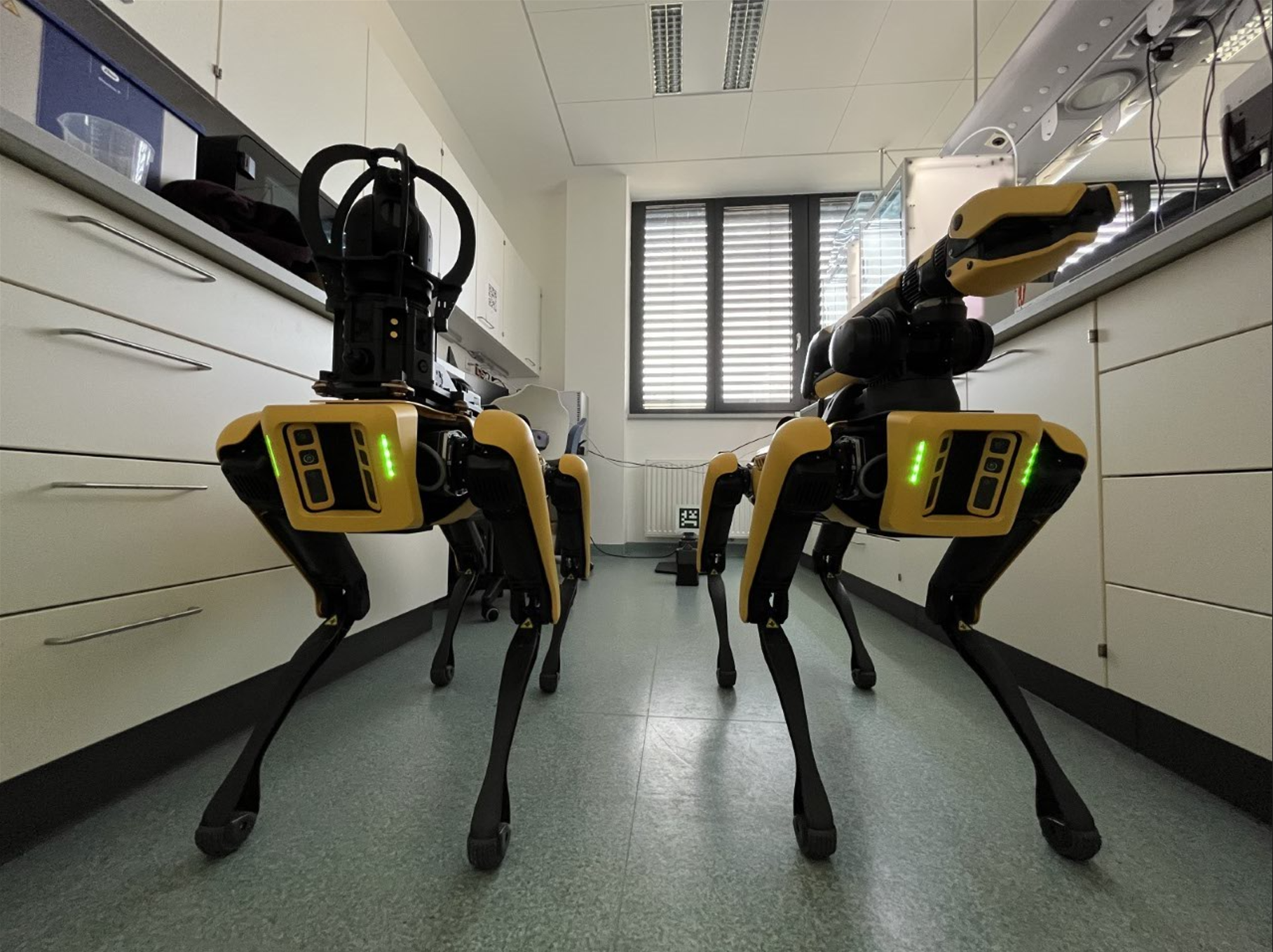}
    \caption{Visual image of the two Spot units utilized for development. Each Spot unit features a different payload configuration. Here, we see the Spot with the pan-tilt-zoom (PTZ) camera (left) and Spot with the arm (right).}
    \label{fig:spot_pack} 
\end{figure}

The paradigm shift towards automating laboratory-based activities also included the integration of increased computing power and robotic platforms; on the one hand, to aid the lab personnel and, on the other, to accommodate the hike in complexity. As such, robots have already been around in laboratories for decades, and the efficacy of the adoption has been well documented \cite{Olsen2012TheScientist}. In recent years, the COVID-19 pandemic further accelerated the adoption of modern robotics and other intelligent technologies in laboratories and clinical environments \cite{khamis_robotics_2021}. 
Traditionally, repetitive tasks like pipetting have been achieved by integrating Cartesian pipetting stations into the lab workflow. Commonly known vendors, such as Tecan or Hamilton, are used for sample preparation, high throughput screening, and execution of bioassays \cite{Kong2012AutomaticArt}.  
A novel advancement towards the beginning of the millennium was the appearance of bench-top sample transportation robots \cite{Thurow2022SystemApplications}. This enabled the feasibility of connecting stand-alone unit operations both physically and by means of a higher-level, over-arching control system. The intricacy of processes called for more versatile solutions, thus leading to the appearance of mobile manipulators (MoMas) in laboratory environments \cite{Abduljalil2019}. This was made possible by the advancements in the fields of mobile- and service robotics, including advanced navigation and motion planning capabilities \cite{Pages,stancel_indoor_2021}. Wheeled MoMas are already becoming ubiquitous for simple pick-and-place sample transportation applications \cite{Kleine-wechelmann2022DesigningLaboratory}. However, some laboratory environments require even more mobility from the robot's side. For example, conventional wheeled MoMas are not capable of opening doors, climbing stairs, or retrieving information via an integrated sensing module. Moreover, use cases, such as remote inspection and error handling, would require robots to support a means of direct remote control or even telemanipulation. Most present-day wheeled MoMas lack the ability to fulfill these applications.

A typical R\&D lab presents a compact environment that is subject to changes in modality and in the requirements of lab operators and engineers. A future state for such an environment would entail robotic platforms which act as lab assistants by either enabling remote presence when an operator is not physically available or as a means of outsourcing easily robotized tasks such as the transport of cargo. The Spot platform presented the most technologically ready and versatile solution compared to the other products available by Boston Dynamics (BD) \cite{SpotDynamics}. 

The aim of this paper is to discuss the initial strategy and path toward the development of Spot as a platform capable of assisting and alleviating the workload that a lab operator has. Moreover, we discuss how Spot overcomes specific limitations seen by present-day MoMa solutions and evaluate areas of improvement in the context of lab automation. For this purpose, we utilize two Spot units, each featuring a different payload configuration, as shown in Fig. \ref{fig:spot_pack}. 


In Section \ref{sec:use_li_sci}, we focus on use cases in life science laboratories. We assess the user needs in the form of a non-representative survey and list a few considerations regarding their implementation. In Section \ref{sec:spot}, we introduce the Spot platform as a means to overcome some of the limitations that state-of-the-art solutions are facing. We first focus on Spot's basic capabilities and typical applications. In Section \ref{sec:pocs}, we present our proof-of-concept (PoC) study on remote inspection. We specifically do this in a life science laboratory setting, which is a new domain, compared to state-of-the-art applications. We also discuss the usability of Spot's manipulator arm and the capacity to execute autonomous missions. In Section \ref{sec:future_work}, we provide an outlook on the anticipated advancements in terms of Spot's capabilities. Finally, in Section \ref{sec:discussion}, we conclude and summarize the paper.

\section{Use cases in life science laboratories}  \label{sec:use_li_sci}

In order to assess potential use cases specific to an R\&D lab footprint, we conducted virtual interviews and a user survey. Both formats began with a high-level introduction to Spot and its out-of-the-box capabilities. The participants included both operators and engineers, as well as project leads and department heads from the process development team for biologics.
The purpose of the interviews and user survey was to leverage the experience of participants familiar with lab-based activities and identify specific areas in which Spot's capabilities can be utilized in order to alleviate workload or provide assistance.  
Results from the virtual interviews and user surveys were clustered based on the application examples, including visual inspection via remote operation, data capture and processing, and robotic manipulations. 

\textbf{One-on-one meetings}  \label{sec:meetings} 
\begin{itemize}
    \item Participants were introduced to Spot via a short presentation and were then asked for feedback on potential use cases for Spot within a lab footprint.
    \item Notes were taken for further evaluation 
\end{itemize}

\vfill\eject

\textbf{Online Survey}  \label{sec:online_surv} 
\begin{itemize}
    \item Participants were provided with the link to a forms-based user feedback survey. 
    \item Results were automatically processed and presented.
\end{itemize}

Results from both the interviews and user surveys confirmed that the remote operation of a robot would provide robust insight into the lab environment in the case of an adverse event. Operators \& engineers referred to scenarios during on-call duty whereby an alarm was triggered by deviation from a set point within a given unit operation. Examples of unit operations include but are not limited to, bench-top bioreactors, surge tanks, and chromatography units. Participants envisioned a scenario where they could sign in and drive the robot to the site of the alarm and utilize an onboard camera to assess the event better. Additionally, participants claimed that they would be better suited to confirm if the alarm was indeed caused by a deviation or a false positive, thus preventing the operator from needing to travel on-site. 
Participants also identified sample transportation as a potential application that could be outsourced to a robotic operator. Additionally, participants mentioned how labor-intensive and time-consuming sample transportation from work cell to work cell can be. 

After being introduced to the robot's capabilities and industrial applications, the majority of participants suggested use cases that could be implemented within the out-of-the-box capabilities. These included remote and autonomous inspection. Participants also envisioned the possibility of utilizing the manipulator arm. Examples include controlling the arm remotely and intervening in the case of errors in automated lab processes. The next release of the Scout interface is expected to make this possible by enabling the control of the manipulator arm.

Another possible use case is the concept of using Spot within an ecosystem of various other autonomous mobile robots (AMR). The general notion would be to have Spot conduct regular reconnaissance missions, checking the lab footprint and ensuring that wheeled mobile robots have a clear path from start to destination. 

Participants from a Process Analytic background identified sample transportation and robotic manipulations as a potential use case. Participants within process analytics usually analyze samples stored in ANSI-SLAS format\footnote{Meets the Standards ANSI/SLAS 1-2004 through ANSI/SLAS 4-2004.} labware such as microtiter plates. Pick and place application would represent the opportunity to alleviate technicians of transporting samples from one work cell to another or from one room to another, an otherwise laborious and time-consuming task. The precise motion control enabled by the robotic arm provides a solution devoid of sequential teaching of robotic vectors. However, the success rate for picking objects of this kind is not optimal for routine sample transportation. Possible workarounds could center around a designated 'sample pick up' station, whereby Spot would navigate towards and complete a predefined sequence of movements that has been already robustly tested. Additionally, a 3D printed stowing section could be designed and mounted onto Spot, allowing the quadruped to pick and stow the samples, ensuring that samples are not exposed to irregular movements due to the nature of the manipulator arm whilst picking up and carrying objects. 

\section{Identifying a suitable mobile robotic platform} \label{sec:spot} 
To answer the need expressed by the users for an easy-to-use remote inspection solution, we reviewed the currently-available mobile robot solutions on the market. Our attention was brought to BD's portfolio.

To counter the effects of labor shortage, increase working efficiency and mitigate monotonous and repetitive tasks, BD is striving towards providing general-purpose robots which will have the ability to serve in human-centric environments. The path envisioned starts with implementing robust mobility and subsequently developing more advanced autonomy. BD has already demonstrated how wheeled and legged robots can achieve the desired level of flexibility and maneuverability. The ultimate goal is to implement skilled and niche manipulation, for which BD has developed the Atlas project as a basic research platform. Atlas is a biped robot designed to perform complex maneuvers and provides insight into how a humanoid robotic platform would handle mundane tasks by interacting with tools and equipment originally designed for humans \cite{Atmeh2014ImplementationRobot}. 

Stretch, on the contrary, is a mobile robotic platform intended for heavy-duty, repetitive tasks typically found in warehouse operations, i.e., on- and off-loading of logistics trucks as well as case handling. Therefore, the Stretch unit features an articulated robotic manipulator arm with an advanced gripper, in addition to advanced computer vision technology for object detection, identification, and sequence planning \cite{Ackerman2022AHour, StretchDynamicsb}. 

Finally, Spot, the quadruped robot, is now a mature commercial product boasting a plethora of versatile capabilities. Equipping it with advanced perception and simple manipulation features is the next step. Our paper provides an overview of the usage of the currently-available functionalities, and we also provide an outlook on what is currently being worked on, specifically focusing on Spot.

\subsection{The Spot quadruped robot}  \label{sec:capa_payl} 
\begin{figure}[ht]
\centering
    \includegraphics[width=\linewidth]{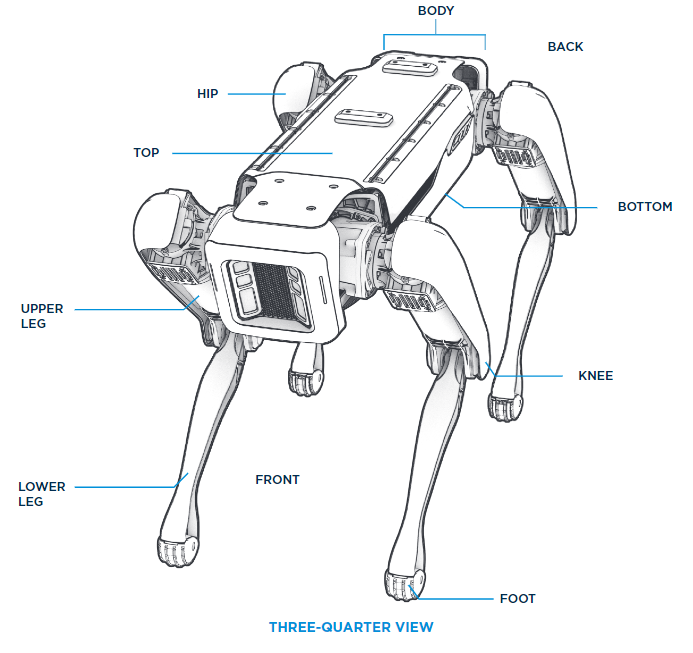}
    \caption{Schematic representation illustrating the key body components of the quadruped. \cite{SpotAnatomy}}
    \label{fig:spot_schematic} 
\end{figure}

Spot's basic capabilities center around its advanced mobile autonomy, which is further enhanced by its quadruped design. Therefore, it is capable of robustly navigating a variety of different surfaces and terrains, including rocky environments as well as stairs, giving it an advantage over wheeled mobile robots. Figure \ref{fig:spot_schematic} illustrates the main body components of the robot that provide this level of autonomy. Additionally, its IP54 rating (ingress protection rating, which assesses how well a device is protected against dust and rain) allows it to operate outdoors, even in rainy or dusty circumstances.

The quadruped features projected stereo cameras facing in five directions, permitting a 360-degree field of view. This enables localization and obstacle detection for autonomous navigation. The robot generates a 3D representation of its surroundings which in turn allows for path planning.

\vfill\eject
Regarding connectivity, Spot uses WiFi to communicate with other devices, such as the standard tablet controller provided by BD. Spot is capable of hosting other clients via its own WiFi access point or as a client, allowing it to be integrated into broader networks for communication with multiple devices. This forms the basis for the networking architecture, which was used to establish the remote troubleshooting use case. 

Additionally, Spot's application can be further enhanced by the integration of multiple different payloads, which are either provided by BD or can be custom-made. A variety is offered by BD \cite{PayloadsDynamics}, including:
\begin{itemize}
    \item A pan-tilt-zoom (PTZ) camera for 360\degree  view, a high-resolution robotic zoom camera, and a thermal \& infrared (IR) detector, enabling image acquisition and sensing (Fig. \ref{fig:spot_pack}, left).
    \item The Spot arm, (Fig. \ref{fig:spot_pack}, right) is utilized for gross manipulations, such as pick and place, constrained manipulation, which entails turning valves, pulling levers, etc., and door opening. 
    \item Additional computational unit called the Spot Core, which enables the implementation of custom applications that run onboard the robot
    \item Extended autonomy payload (EAP2), which provides a wider range of perception than the surround cameras thanks to a LiDAR sensor. This gives the robot more confidence in localizing itself and navigating in different environments. 
\end{itemize} 

\vfill\eject
\subsection{Typical application areas} \label{sec:appl_ar} 
Typical application areas for Spot include remote-controlled or autonomous operation within industrial settings \cite{YourJobs, QuadrupedGlobalSpec}, i.e., oil rigs, (nuclear) power plants \cite{RadiationPlatform}, construction sites \cite{TheSites, Halder2023ConstructionStudy}, and mines. Disaster and safety response scenarios are also covered, generally, in situations whereby dangerous environments could potentially endanger the lives of human operators.

Usage of Spot and similar quadrupeds also gained prominence and notoriety during the height of the COVID-19 pandemic. The Unitree Laikago quadruped was utilized for enforcing social distancing \cite{Chen2021AutonomousRobot}. Spot was used within healthcare and clinical settings in order to minimize and mitigate risk and exposure between healthcare workers and patients \cite{BostonDynamicsHealthcareResponse}. Spot's custom payload integration enabled telepresence, delivery of medicine to patients, and inspection of patients' vital signs; in particular, body temperature, which was a key indicator of possible infection. 

Spot's utilization within healthcare and clinical settings opens the door for potential future states whereby mobile robots of a similar framework are used for a variety of applications, two examples that would provide benefit to a wide range of medical or pharmaceutical environments are disinfection of either a hospital room or a lab with a high biosafety level (BSL) and internal delivery of particular goods within a given environment. 
Life science laboratories, particularly research and development (R\&D), require a different set of requirements, which we have begun to test and evaluate.

\section{Considerations}  \label{sec:considerations} 
The strengths of Spot and its out-of-the-box capabilities lie in navigating different terrains and having the ability to be equipped with different payloads for different tasks. The typical lab robot use case of pick-and-place labware transportation is out of scope for now. Precise motion planning can be achieved through the Spot API and will be evaluated further in future experiments. Limitations of the robotic arm module mainly center around the end effector, which is currently not suited towards handling labware such as microtiter and deep-well plates, typically used in an R\&D lab. It would be possible to create a custom end effector specific for handling labware of this nature.
The standing height of the robot is also a limitation, as it is unsuitable for a wide coverage of bench-top devices and unit operations. Moreover, the standing height of the robot and the range of the Spot arm also impact and limit the set of robotic manipulations that it can conduct.

\section{Proof-of-concept studies}  \label{sec:pocs} 
\subsection{Equipment}  \label{sec:equipment} 
In order to conduct the PoC studies, we used the two Spot units introduced in Section \ref{sec:intro}. Both were equipped with the computational unit, called Spot core, which enables advanced data processing and enhances communication applications onboard the robot. One of the units was also fitted with the robotic manipulator arm, called Spot Arm, whereas the other unit carried the PTZ camera as the primary payload. 

\begin{figure}[ht]
\centering
    \includegraphics[width=\linewidth]{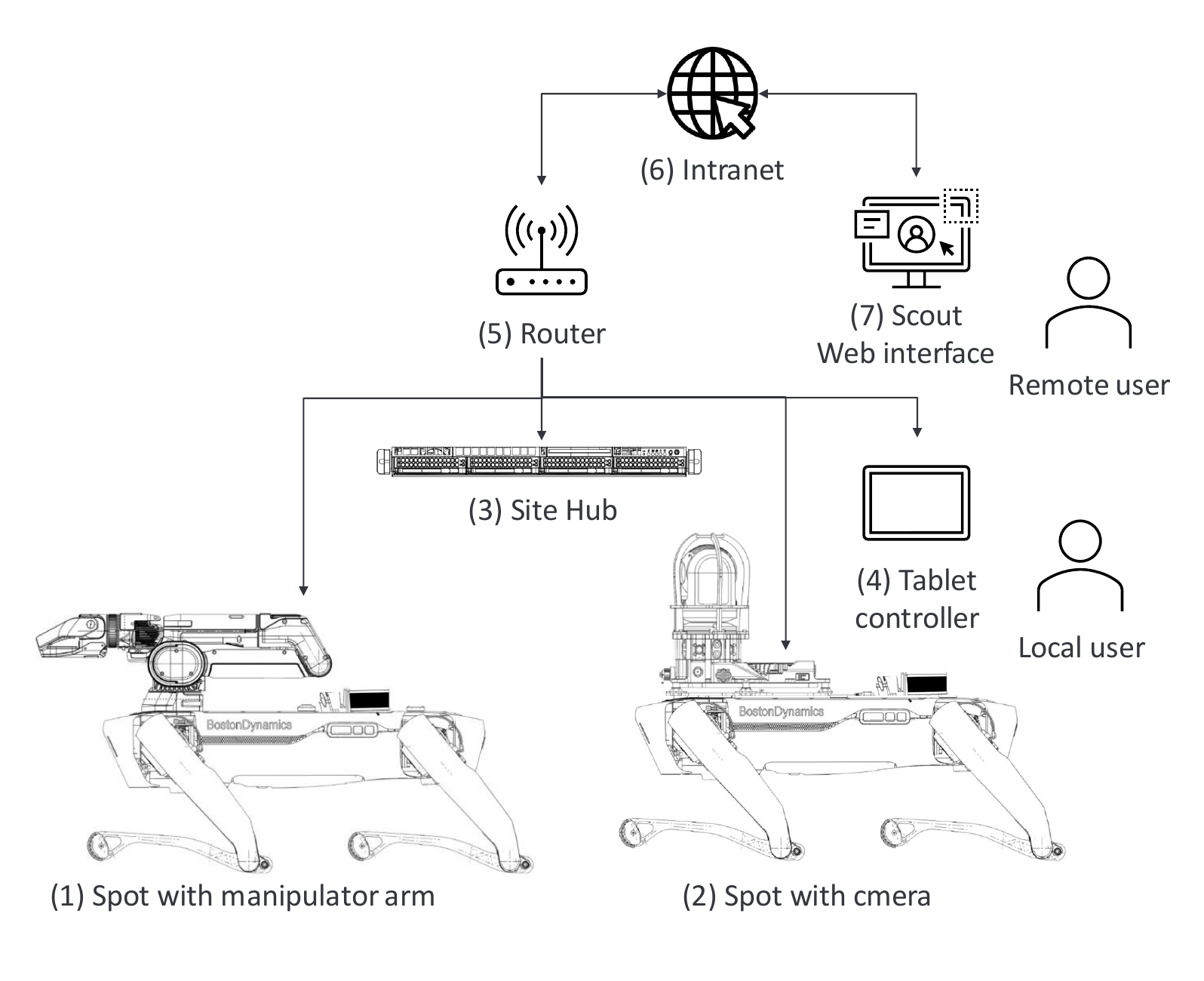}
    \caption{System architecture of the experimental network setup. Spot units (1) and (2), equipped with a Spot Arm and a PTZ camera, respectively; Local server unit (3) Site Hub, hosting the Scout interface; Local tablet controller (4) for configuration, mission creation, and human-initiated autonomy; Local router (5) and access point; Global Intranet (6); Remote web browser displaying the Scout interface (7)}
    \label{fig:network} 
\end{figure}

Enabling remote access was a key factor in setting up the experimental environment. This involved connecting both robots to a wireless local area network (W-LAN) in client mode. The Site hub server, which hosts the web-based control interface, needs to be integrated into the corporate network. The robots must be in the same domain, reachable (routed) from the Site Hub and the control tablet. Thus, within the corporate network (intranet), any user laptop would be able to reach the browser interface. Fig. \ref{fig:network} shows the experimental network setup.

\subsection{Remote inspection}  \label{sec:remote_insp} 
We conducted PoC studies within a pharmaceutical laboratory, which was geared toward process development. We utilized the Spot with the PTZ camera, in addition to the Spot with the robotic arm and the web control interface known as Scout. 
Primarily focusing on the remote inspection use case, specifically visual inspection, we introduced Spot as a robotic platform and presented its out-of-the-box capabilities in order to give the participants a high-level overview. Thereafter, we clustered the use cases stemming from the user surveys and interviews into specific applications.
We focused on a scenario whereby a device had lost connection to a process control system or if it were an outdated (legacy) device without a suitable interface and would require a readout of a display. 

This enabled us to perform the tests in the aforementioned sections. This included the following steps:
\begin{itemize}
    \item Logging in to the Scout interface from the office PC and activating the robot
    \item Driving it via the camera views and the 3D representation of the environment to the designated laboratory
    \item Switching to Inspection mode; activating the PTZ camera
    \item Localizing the instrument for the readout
    \item Zooming in and triggering an image acquisition, thus storing an image locally
    \item Navigating back to the idle position of the robot in the previous room, where the experiment finished.
\end{itemize}

\begin{figure*}[]
    \centering
  \subfloat[Driving with the 360\degree camera.\label{fig:driving_cam}]{%
       \includegraphics[width=0.455\linewidth]{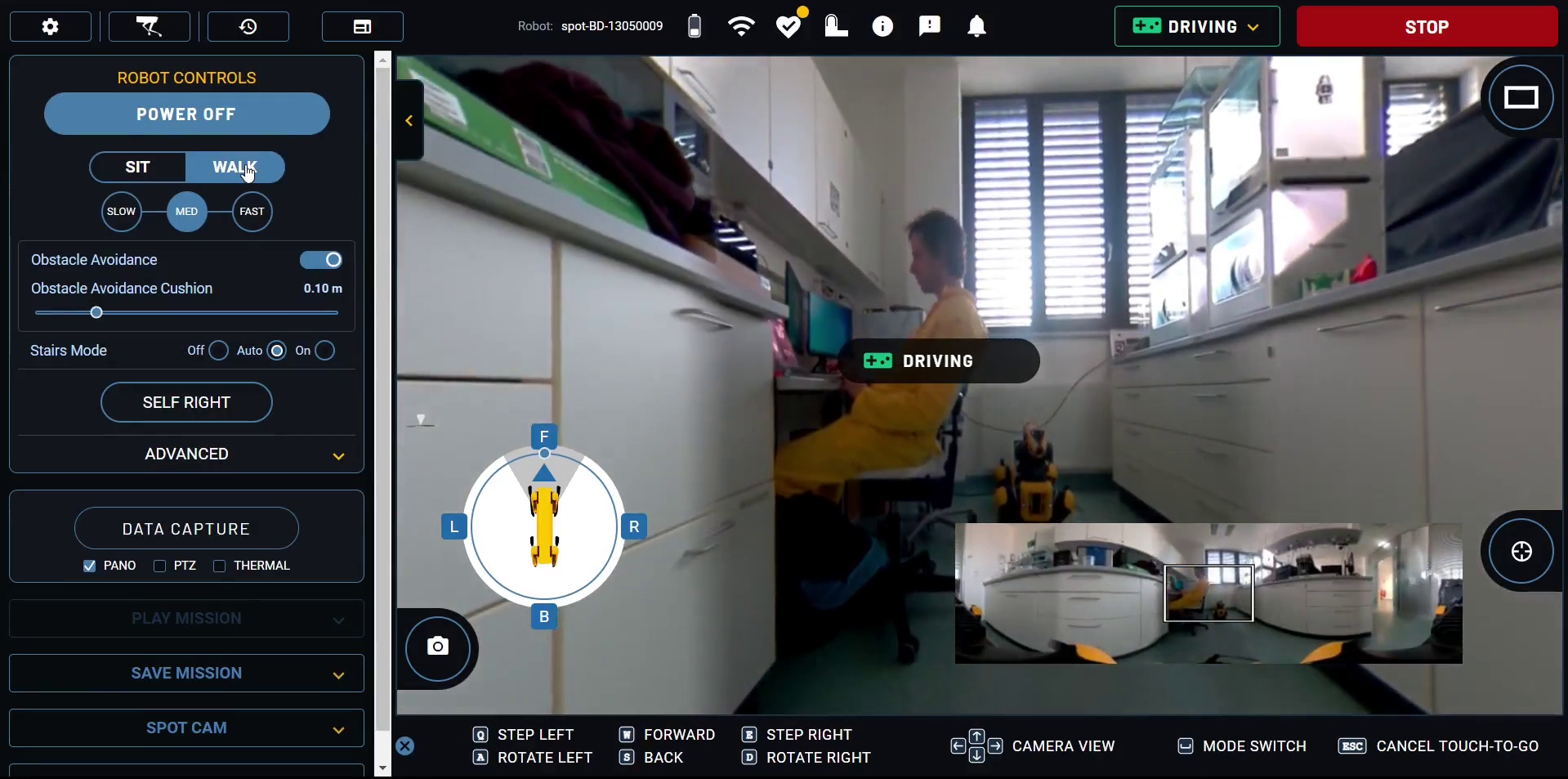}}
    \hspace{7mm}
  \subfloat[Driving with the 3D environment representation\label{fig:driving_3D}]{%
        \includegraphics[width=0.45\linewidth]{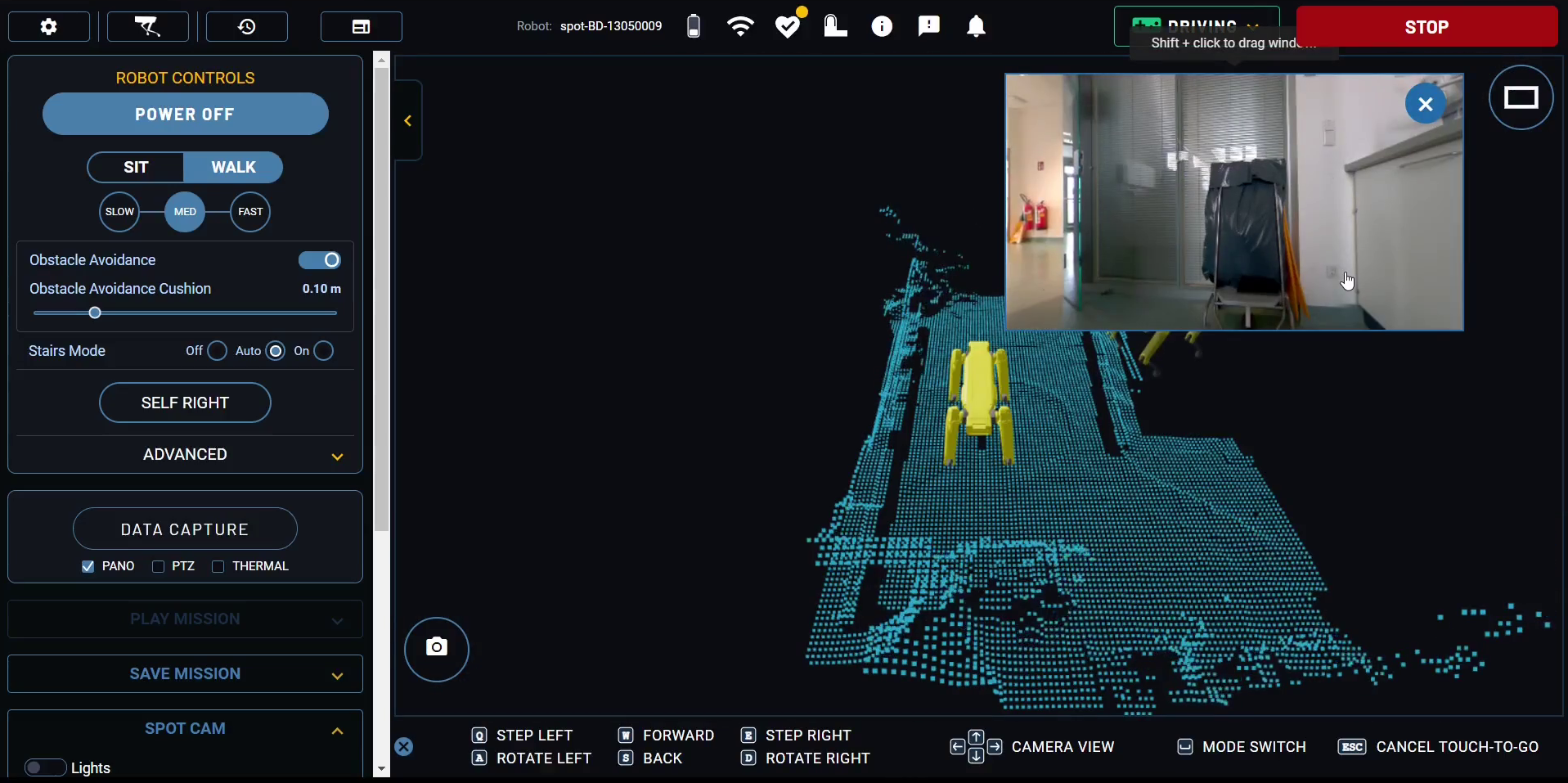}} \\
        
  \subfloat[PTZ camera view of the laboratory bench \label{fig:ptz_bench}]{%
       \includegraphics[width=0.455\linewidth]{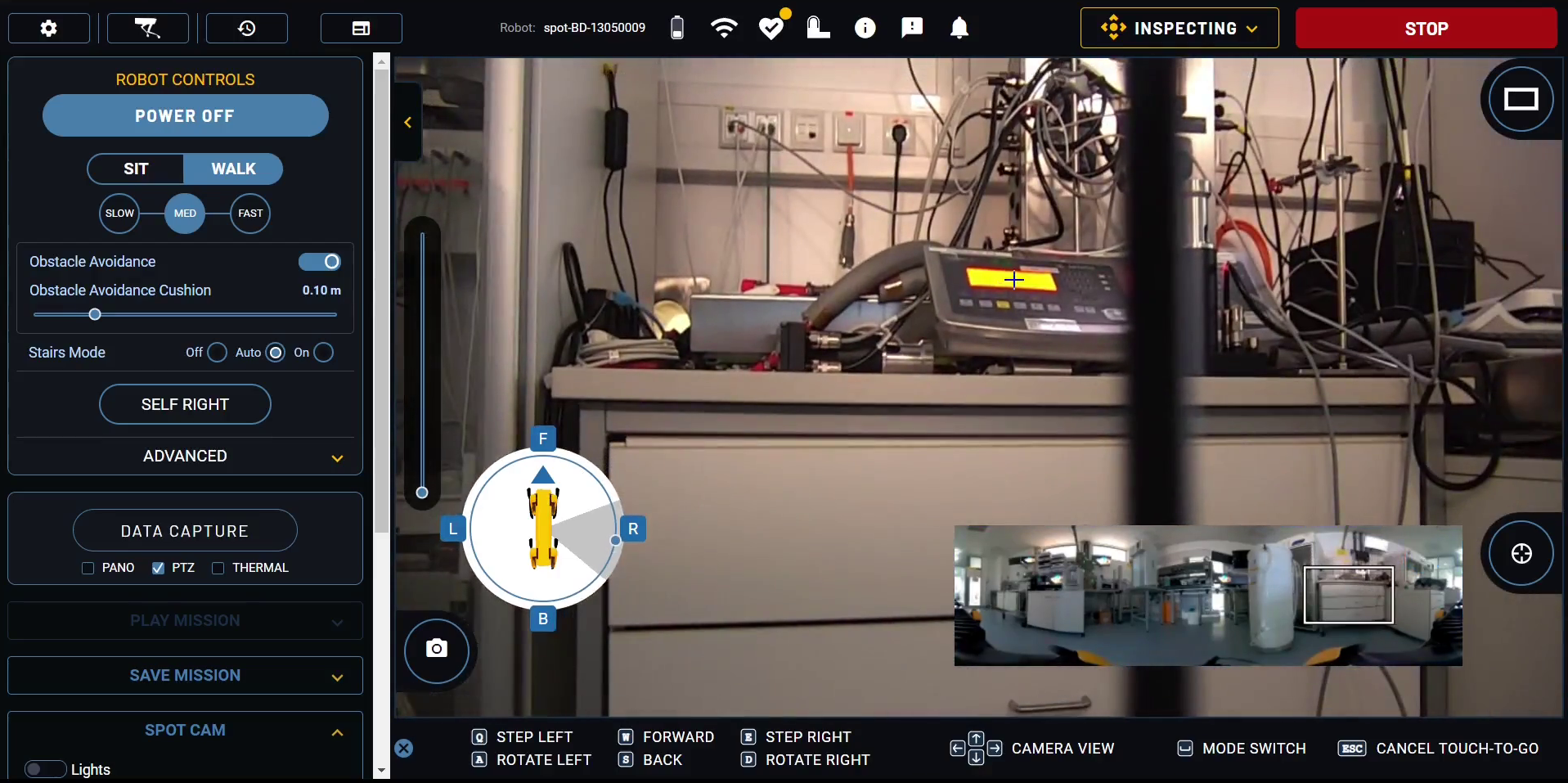}}
    \hspace{7mm}
  \subfloat[Zoomed-in view of the scale display\label{fig:ptz_display}]{%
        \includegraphics[width=0.45\linewidth]{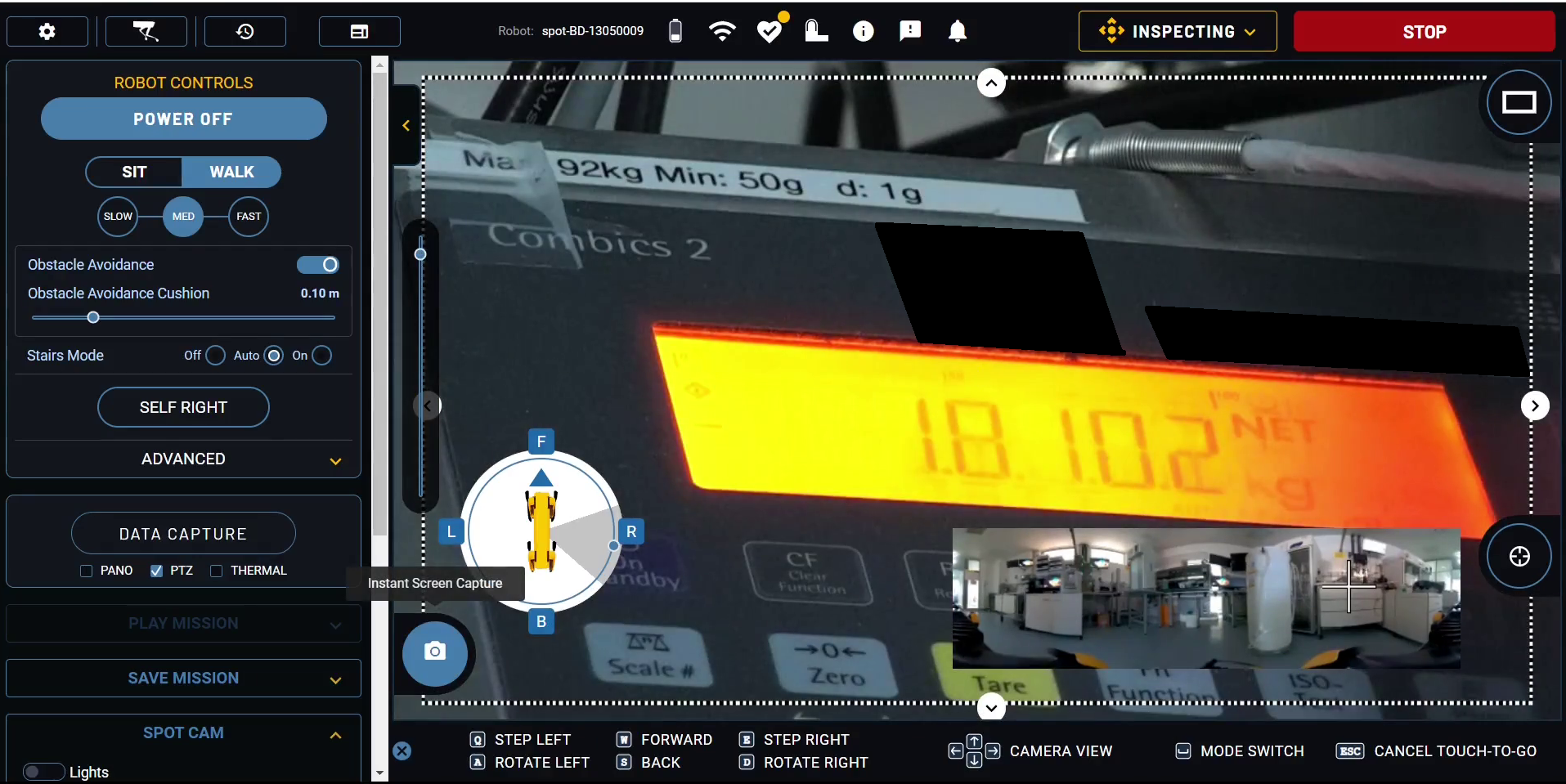}} 
        
  \caption{Series of screenshots from the Scout solution control center used to conduct the remote troubleshooting use case.}
  \label{fig:scout_experiment}
\end{figure*}

Fig. \ref{fig:scout_experiment} shows four key screenshots of the experiment. On subfig. \ref{fig:driving_cam}, the main view of the PTZ-camera can be seen while on subfig. \ref{fig:driving_3D}, the real-time pointcloud representation of the environment is shown. The target instrument can be seen in subfig. \ref{fig:ptz_bench}, placed on a conventional laboratory bench. Finally, subfig. \ref{fig:ptz_display} shows the readout image from the scale's display. The low viewing angle results in a sub-optimal view on the LCD screen, resulting in a slight ghost effect. The height of Spot is a limiting factor, both in the case of inspection activities in human environments and in terms of bench-top manipulation.

\subsection{Manipulation}  \label{sec:manip} 

Spot Arm is capable of user-initiated autonomy, in which the user triggers certain actions via the controller interface enabled by the Android tablet. Users can select specific objects for the robot to pick by point and click. It should be noted that the user has to keep cautious that the robot arm does not collide with the environment. Additionally, implementing robotic manipulations within pre-recorded missions (otherwise known as auto walk) is also a possibility.

Currently, control of the manipulator arm is not available via the Scout solution. This is a feature that BD is currently developing and will address with a future release.

\section{User tests}

We conducted user tests aimed at simulating a scenario whereby operators remotely access, drive \& return the Spot robot through a given lab footprint, identify PoIs that would be subject to technical difficulties and conduct a visual inspection.
The user test began by giving participants an introduction to the Scout Solution interface, followed by a demonstration on how to remotely control and navigate the Spot and operation of the Spot CAM payload. 
Users were then asked to remotely control and navigate the robot from a home position to an upstream process development lab and visually inspect unit operations such as Bioreactors and their Direct control units (DCU). Following visual inspection and visual data captures, participants then returned Spot to the home position and docked the Spot unit. 
After the practical section of the user test had concluded, participants were asked to complete a questionnaire, where they rated the user interface and user experience and provided feedback. 
A total of 9 participants took part in the user test. Participants were from various technical and non-technical functions within the R\&D organization and had varying experience levels in working and controlling robots. 
Most correspondents reported that the overall user interface and experience were very intuitive and provided a relatively easy means of controlling the robot, despite it being their first time operating a robot of any kind remotely. On the other hand, participants also mentioned experiencing latency whilst remotely controlling Spot \& the Spot CAM payload, which resulted in poor response time.  
The users, on average, rated the remote control function at 6,4 (out of 10) and the user interface and user experience at 7,2 (also out of 10).


\section{Discussion}  \label{sec:discussion} 
Operators \& technicians are required to respond to alarms during off-duty hours during process development. Examples can include PO2 \& CO2 sensor alarms for Bioreactor unit operations, and other examples include alarms detecting a significant change in temperature for laboratory refrigerators, housing equipment such as filtration units, surge tanks, or chromatography units.
If we consider the operators and engineers that are working on continuous and advanced process development that live in close proximity (10-20km) to the facility, this would translate to approximately 20 minutes (best case scenario) of travel time via car, followed by an additional 15-20 minutes to: pass security, wear appropriate lab personal protective equipment (PPE), check the status of the alarm, and take adequate measures to assess and then mitigate the problem. Moreover, alarms and sensors are subject to producing false positive alarms, and the arbitrary information provided by DCUs for specific unit operations may not adequately explain the exact root cause, i.e., issues caused by sub-components such as valves or tubing. 
Utilizing the Scout solution as a first step could save 30-40 mins of travel and operation time by navigating Spot to the PoI and assessing the issue visually via remote presence. 
A future perspective would involve Spot being employed as a first line of action against alarms that are caused by faulty gas sensors\& alarms. This has wider implications in terms of its deployment seeing as this is a topic that affects all site labs and aligns with the objectives of the environmental, health, and safety (EHS) department. 

\section{Outlook and future work}  \label{sec:future_work} 
Part of the remote inspection activity could be implemented with the autonomous navigation functionality, the Autowalk, in that the user would not need to drive the robot manually but record a specific path. Then, Spot could perform this pre-programmed path to arrive at a Point Of Interest (PoI). However, to achieve full autonomy, the enhanced autonomy module (EAP2) would be needed, and topological mapping needs to be implemented to enable arbitrary navigation between PoI's in a graph-like representation. Oxford Robotics Institute's (ORI) AutoInspect solution offers Simultaneous Localization and Mapping (SLAM) of this sort \cite{Ramezani2020OnlineClosure}. 

Human-initiated autonomy is a key principle in leveraging mobility, perception, and manipulation skills. This represents a middle ground between full autonomy and direct telemanipulation. The former means that the robot performs pre-recorded actions and uses built-in algorithms for decision-making. This includes high-level semantic planning, whereby a set of actions is laid out based on a desired state specification. The actions then need to be translated to low-level physical interactions with the environment, e.g., in the form of motions being expressed geometrically \cite{Leidner2012ThingsClasses}. In the case of telemanipulation, a human is responsible for determining the sequence of actions and also of executing the motions. The robot, in this case, can be considered to be an extended actuator for the human brain. In most cases, the motions of the human operator are mapped to the robotic actuators. A typical example is surgical robotics, where the surgeon controls a special master joystick, and the movements are executed by one or more endoscopic actuators \cite{Ngu2017TheSurgery}.

In this case, delays between the master (controller) and the slave (actor) units' motions can be catastrophic. In most instances, however, there is a direct-wired connection, and the surgeon is sitting in the next room. In situations whereby the signal has to travel greater distances, the delays can impede the usability of the system. A technique to resolve this issue is to leave a certain amount of autonomy to the robotic actor, as presented by Schmaus et al. in the context of orbit-to-ground teleoperation for space exploration applications \cite{Schmaus2020KnowledgeCoworker}. By automating certain sub-tasks and leaving only high-level control to the (remote) human operator, the time-critical actions can be controlled locally, thus enabling a quicker response. Quere et al. present the concept of shared control templates for assistive robotics \cite{Quere2020SharedRobotics}. Key factors to enable this are suitable knowledge representation frameworks \cite{Leidner2012ThingsClasses} and appropriate cognition on the robot's side, which enables autonomous task planning and execution.

Potential future work may include implementing the above-mentioned concepts for Spot and/or other mobile manipulator platforms. This will include further assessing BD's present and upcoming capabilities regarding manipulation. Furthermore, the perspective of combining advanced telemanipulation technologies, such as SRI's Robot Telemanipulation System \cite{TelemanipulationInternational} with different autonomous mobile robot (AMR) platforms, needs to be evaluated.

\section{Conclusion}  \label{sec:conclusion} 
We provided a perspective on utilizing Spot, the quadruped robot of BD, in a pharmaceutical R\&D laboratory environment. We assessed the capabilities of two specific payloads, including the pan-tilt-zoom camera and the manipulator arm. We conducted user interviews to generate use cases that are specific to a lab footprint. We did initial testing in the scope of remote troubleshooting with BD's web-based Scout interface. Finally, we provide an outlook on manipulation and other advanced capabilities.

\section*{Acknowledgements}
This work was funded by Baxalta Innovations GmbH, a Takeda company. We thank our teammates for the fruitful collaboration and their help in reviewing the article:  Michael Schwaerzler, Masatoshi Karashima, Patricia Wildberger, Nozomi Ogawa, and Seishiro Sawamura.

This work was supported by the Doctoral School of Applied Informatics and Applied Mathematics, Óbuda University.

Péter Galambos and Károly Széll thankfully acknowledge the financial support of this work by project no. 2019-1.3.1-KK-2019-00007 implemented with the support provided from the National Research, Development and Innovation Fund of Hungary, financed under the 2019-1.3.1-KK funding scheme. Péter Galambos is a Bolyai Fellow of the Hungarian Academy of Sciences. Péter Galambos is supported by the UNKP-22-5 (Bolyai+) New National Excellence Program of the Ministry for Innovation and Technology from the source of the National Research, Development and Innovation Fund.

We thank Boston Dynamics for reviewing and approving the article.

\section*{Conflict of interest statement}
Brian Parkinson and Ádám Wolf are employees of Baxalta Innovations GmbH, a Takeda company, Vienna,
Austria.

\bibliographystyle{IEEEtran}
\IEEEtriggeratref{14}
\bibliography{references,acta}
\end{document}